\documentclass[transaction]{IEEEtran}
\usepackage{times}
\usepackage{fancyhdr,graphicx,amsmath,amssymb}
\usepackage{algorithm}
\usepackage{algpseudocode}
\usepackage{graphicx}
\usepackage{enumitem}
\usepackage{subcaption}
\usepackage{multirow}
\usepackage{amsmath}
\usepackage{mathtools}
\usepackage{balance}
\usepackage{url}
\usepackage{booktabs}
\usepackage{breqn}
\usepackage{amsfonts,amssymb}
\usepackage[noadjust]{cite}
\usepackage{color}

\begin{document}

\title{MDPose: Human Skeletal Motion Reconstruction Using WiFi Micro-Doppler Signatures}

\author{Chong~Tang,
        Wenda~Li,~\IEEEmembership{Member,~IEEE,}
        Shelly~Vishwakarma,
        Fangzhan~Shi,
        Simon~Julier,~\IEEEmembership{Member,~IEEE,}
        and~Kevin~Chetty,~\IEEEmembership{Member,~IEEE,}
\thanks{C. Tang, W. Li, S. Vishwakarma, F. Shi, and K. Chetty are with the Department of Security and Crime Science, University College London, UK e-mail: (chong.tang.18@ucl.ac.uk, wenda.li@ucl.ac.uk, s.vishwakarma@ucl.ac.uk, fangzhan.shi.17@ucl.ac.uk, k.chetty@ucl.ac.uk).}
\thanks{S. Julier is with the Department of Computer Science, University College London, UK email: (s.julier@ucl.ac.uk)}
}

\maketitle

\begin{abstract}
Motion tracking systems based on optical sensors typically often suffer from issues, such as poor lighting conditions, occlusion, limited coverage, and may raise privacy concerns. More recently, radio frequency (RF)-based approaches using commercial WiFi devices have emerged which offer low-cost ubiquitous sensing whilst preserving privacy. However, the output of an RF sensing system, such as Range-Doppler spectrograms, cannot represent human motion intuitively and usually requires further processing. In this study, MDPose, a novel framework for human skeletal motion reconstruction based on WiFi micro-Doppler signatures, is proposed. It provides an effective solution to track human activities by reconstructing a skeleton model with 17 key points, which can assist with the interpretation of conventional RF sensing outputs in a more understandable way. Specifically, MDPose has various incremental stages to gradually address a series of challenges: First, a denoising algorithm is implemented to remove any unwanted noise that may affect the feature extraction and enhance weak Doppler signatures. Secondly, the convolutional neural network (CNN)-recurrent neural network (RNN) architecture is applied to learn temporal-spatial dependency from clean micro-Doppler signatures and restore key points' velocity information. Finally, a pose optimising mechanism is employed to estimate the initial state of the skeleton and to limit the increase of error. We have conducted comprehensive tests in a variety of environments using numerous subjects with a single receiver radar system to demonstrate the performance of MDPose, and report $29.4mm$ mean absolute error over all key points positions, which outperforms state-of-the-art RF-based pose estimation systems.

\end{abstract}


\IEEEpeerreviewmaketitle

\section{Introduction}
\label{sec: Introduction}
\IEEEPARstart{W}{ith} the rapid development of the Internet of Things (IoTs) and smart buildings in recent years, users can engage with their surroundings in a more natural way, such as through voice and pose. This requires accurately tracking human movements and interpreting them properly. One of the most promising solutions is based on computer vision and machine learning algorithms\cite{zhang2012rgb, ristani2018features}. However, camera based systems suffer from issues around lighting conditions, occlusion, coverage and privacy. This motivates the emergence of RF-based techniques using ubiquitous WiFi signals which offer low-cost precise sensing in a variety of scenarios whilst preserving privacy. 

Many WiFi-based sensing methods use the amplitude changes of the channel state information (CSI) to analyze the human motion properties, and have validated their feasibility in the fields of indoor localisation\cite{gao2017csi}, activity recognition\cite{wang2017device, wang2015understanding} and healthcare\cite{tan2018exploiting}. However, few CSI-based systems can use the phase information due to unsynchronised local oscillators in WiFi networks\cite{tewes2021ws}, resulting in a loss of the important Doppler information. More recently, the passive WiFi radar (PWR) system provides an alternative for the WiFi-based sensing tasks. There are many applications that have been successfully demonstrated by PWR systems, including occupancy detection, activity classification and hand gesture recognition\cite{tang2020occupancy, li2018passive, gurbuz2020radar, palama2021measurements}. Different from the CSI-based systems, PWR can express not only the amplitude of returned signals, but also the micro-Doppler frequency shifts in different body parts using micro-Doppler ($\mu$-Doppler) spectrograms, which offers a more comprehensive perception of human movements. Furthermore, it is a passive radar system in which the signal illumination is independent of the radar itself. This implies that no modifications to the existing WiFi network are required, instead the CSI-based systems still need additional network interface cards. Therefore, PWR has the great potential to improve the performance of the current WiFi-based sensing.

However, neither $\mu$-Doppler spectrograms nor CSI amplitudes are intuitive to ordinary users. Even if some general information, such as human activity categories, can be obtained using machine learning algorithms, more details of the motion are still not available. In this case, there is considerable interest in wireless sensing field which aims to reconstruct more detailed human skeletal motions from RF signals. Zhao et al.~\cite{zhao2018through, zhao2018rf} proposed RF-Pose and RF-Pose3D to achieve human skeleton estimation with only RF signals. The frameworks used RGB frames or motion capture (Mocap) data as the supervision to train CNN-based skeleton reconstruction model. From results, we can see that with the skeletal estimation, wireless sensing systems present comparable performance to traditional camera-based systems and outperform them in extreme conditions, such as through-the-wall, occlusion and dark environments. However, the RF source they use is a frequency modulated continuous wave with a $1.78GHz$ sweep bandwidth, which requires the construction of additional hardware equipment and may limit the application of the system in some cases. Jiang et al.~\cite{jiang2020towards} proposed the WiPose framework to achieve 3D pose estimation using WiFi CSI information. Their model is based on the CNN-RNN architecture and can recursively calculate poses from the estimated quaternion information. The final results demonstrated that WiFi signals carry sufficient information for human skeleton reconstruction and can also tackle extreme sensing environments. However, as previously stated, the lack of phase information may impede further developments of CSI-based systems. Furthermore, how to initialise the first pose and deal with the accumulation of errors in long-term estimation have not been fully discussed in \cite{jiang2020towards}. Therefore, the system still has potential for improvement.

\begin{figure}
\centering
\includegraphics[scale=0.5]{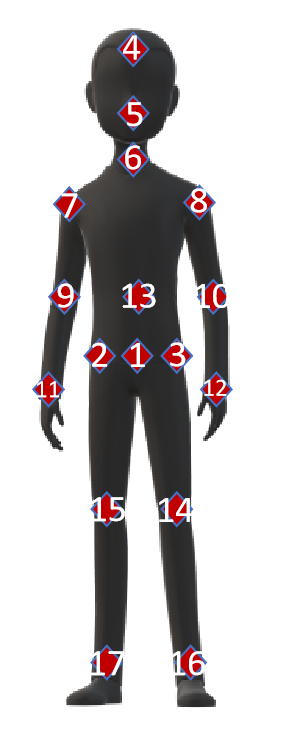}
\caption{The key points locations on human body}
\label{fig: markers}
\end{figure}

The pioneering efforts by researchers in this area has inspired us to develop MDPose, a novel framework for human skeletal motion reconstruction using the commercial WiFi network. After removing noise from measured spectrograms, MDPose can extract velocity information from $\mu$-Doppler spectrograms for up to $17$ skeletal key points (as shown in Fig. \ref{fig: markers}) and recursively calculate poses. Furthermore, MDPose has a pose optimising mechanism that can obtain optimisation vectors based on the current pose and the future's motion to address issues related to the initial pose estimation and the long-term error accumulation. In particular, MDPose enables training the velocity estimation network with simulated spectrograms due to our denoising strategy, which can effectively address the training difficulties caused by the lack of measurement data. Finally, various experiments were carried out in different environments and with different subjects, successfully demonstrating the effectiveness and generalisation performance of MDPose.

The rest content of our paper is organized as follows: First, Section \ref{sec: Framework Overview} briefly describes the framework. Next, Sections \ref{sec: Noise} to \ref{sec: self-optimization} introduce details of three main components of MDPose. Then Section \ref{sec: data collection} explains the dataset, hardware and software we used in experiments, and the training settings of neural networks. Finally, we present experimental results and important discussions in Section \ref{sec: results}, and conclude the work in Section \ref{sec: conclusion}.

\section{System Overview}
\label{sec: Framework Overview}
MDPose is based on $\mu$-Doppler information and allows the reconstruction of human skeletal movements without any modifications of the existing WiFi network. To address a series of challenges in practical applications, we design a novel framework to handle them step by step.
\begin{itemize}
    \item \textbf{Data Collection:} To collect $\mu$-Doppler changes, MDPose uses the PWR system in which the signal illumination i.e. WiFi access point is independent of the radar itself. This implies that no modifications to the existing WiFi network are required. Compared to CSI systems that operate using NIC cards, it can be more easily adapted to a wide range of scenarios.
    \item \textbf{Data Denoising:} Interference and noise may mask important signals and thus affect the extraction of dynamic characteristics. Therefore, the data denoising is the first step in MDPose. In this step, a deep learning-based denoising network is implemented to effectively remove unwanted signals. 
    \item \textbf{Velocity Estimation:} There are many choices to describe the motion property, such as the position, quaternion, etc. For the $\mu$-Doppler, it is strongly related to velocity profiles. Therefore, in our case, estimating velocity could have more accurate results compared with using other types of data. So, we propose a CNN-RNN neural network to extract velocity information. 
    \item \textbf{Pose Optimising:} Although we can obtain the velocities from the previous step, we do not know the initial pose $p_0$ so that we cannot calculate the following poses. In this case, we propose a pose optimising mechanism to help estimate a proper $p_0$. Then with $p_0$ and velocities, we can finally calculate the following poses based on the relationship between the velocity and the position. However, unlike the quaternion-based approach\cite{jiang2020towards}, there is no strong constraint between the velocities of the different joints, and therefore unrealistic skeletons are likely to occur as time increases. To address the issue of accumulative error, the pose optimising method can be used again, which can recursively optimise the current pose according to the velocities and poses afterwards.
\end{itemize}
The overall framework of MDPose has been presented in Fig. \ref{fig: MDPose}, and more details will be discussed in the following sections.

\section{Clean Up Micro-Doppler Spectrogram}
\label{sec: Noise}
PWR includes separately located reference and surveillance channels. The reference channel captures the source signals, while the surveillance channels gathers signals reflected off targets but may also receives reflections from environmental clutters, etc. In this section, we introduce the signal processing methods we used to extract the desired $\mu$-Doppler spectrogram, and introduce useful denoising algorithms to clean up the measured data.
\subsection{PWR Signal Processing}
\label{subsec: pwrsp}
If we denote the source signal as $u(t)$, then the signal received by reference channel $S_{ref}(t)$ can be expressed as:
\begin{dmath}
\label{eq: ref}
    S_{ref}(t)=A_{ref}u(t-\tau_{ref})
\end{dmath}
where $S_{ref}(t)$ is the copy of $u(t)$ with a delay $\tau_{ref}$ and amplitude scaling factor $A_{ref}$. For the surveillance channel, its received signal $S_{sur}$ can be expressed as:
\begin{dmath}
\label{eq: sur}
    S_{sur}(t)=\sum_{l=1}^{N}A_{l}u(t-\tau_{l})e^{j2\pi f_{l}t}+\sum_{m=1}^{M}A_{m}u(t-\tau_{m})e^{j2\pi f_{m}t}+A_{DSI}u(t-\tau_{DSI})+\sum_{n=1}^{N}A_{n}u(t-\tau_{n})
\end{dmath}
where $A$ and $\tau$ still are the amplitude scaling factor and the delay that belong to different returns. And we will have four terms that represent the reflected signals from the $L$ targets, the multipath reflection, the direct signal interference (DSI) and the returns from stationary clutter objects in the surrounding environment, respectively. Specifically, we also have Doppler shift $f_{l}$ caused by the movement of targets which is also related to the target velocity profiles, and another Doppler shift $f_{m}$ due to the multipath effect. Next, we can extract $\mu$-Doppler and range information by cross-ambiguity function (CAF) as shown in the following:
\begin{dmath}
\label{eq: CAF}
    CAF(\hat{\tau},\hat{f_{D}})=\int_{-\infty}^{\infty}S_{sur}(t)S_{ref}^{*}(t-\hat{\tau})e^{-j2\pi \hat{f_{D}t}}dt
\end{dmath}
where $\hat{\tau}$ and $\hat{f_{D}}$ are the expected target delay and Doppler shift, respectively, and the $[*]$ is the complex conjugate. This will output CAF results as:
\begin{dmath}
\label{eq: CLEAN}
    CAF(\hat{\tau},\hat{f_{D}})=\sum_{l=1}^{N}CAF_{tgt}^{l}+\sum_{m=1}^{M}CAF_{multipath}^{m}+CAF_{DSI}+\sum_{n=1}^{N}CAF_{clutter}^{n}
\end{dmath}
In Equation \ref{eq: CLEAN}, we can see that the result does not only contain the desired target Dopper-range information, but also have noise caused by multipath reflections, DSI and environmental clutters. Among them, the DSI has the greatest impact on the results, which may mask reflected signals from targets and cause unwanted peaks at the zero-Doppler frequency. To eliminate it, we then apply the CLEAN algorithm as presented in the below:
\begin{dmath}
    CAF^{'}(\hat{\tau}, \hat{f_{D}})=CAF(\hat{\tau}, \hat{f_{D}})-\alpha CAF_{self}(\hat{\tau}-T_k, \hat{f_{D}})
\end{dmath}
where $CAF_{self}(\hat{\tau}-T_k, \hat{f_{D}})$ is the CAF over the reference channel and $\alpha$ is the maximum absolute value of $CAF(\hat{\tau}, \hat{f_{D}})$. Due to the insufficient range resolution, after the CLEAN process, the PWR system only takes $\mu$-Doppler information i.e. velocity profiles from the CAFs and combines them along time axis to generate the final $\mu$-Doppler spectrogram. However, even the DSI can be removed, the multipath noise and the attenuation of the signal with distance may still affect the quality of the spectrogram. Therefore, other effective denoising algorithms are required to further clean up the data.
\subsection{Latent Feature-wise Mapping Network}
\label{subsec: FMNet}
The latent feature-wise mapping network (FMNet) proposed in our work\cite{tang2021fmnet} combines strengths of variational auto-encoder and adversarial auto-encoder that maps noisy measurement latent features to a clean space to achieve the noise reduction. In short, FMNet will initially learn how the various motion attributes are spread over a clean space and how to restore data from the clean space. Given a noisy measured data, it can then find the latent features in this space that have the closest motion information to it. After that, the decoder structure of FMNet will restore the $\mu$-Doppler spectrogram based on the found clean features, and the interference from environmental clutters and multipath noise can be effectively removed. In \cite{tang2021fmnet}, we have demonstrated that compared with other networks\cite{kingma2013auto,makhzani2015adversarial}, FMNet has better generalization ability, up to $10\%$ improvement, which still performs well in some challenging scenarios, such as through-the-wall sensing.

Furthermore, to construct the clean space, we used the simulation data matched to the measurement data to train the FMNet, so the final outputs are simulation-like. More details about FMNet can be found in \cite{tang2021fmnet}.

\section{Velocity and Pose Estimation}
\label{sec: Velocity}
To reconstruct the skeletal motion, an intuitive solution might be directly learning positions of each key points from a $\mu$-Doppler spectrogram. However, common WiFi bandwidth is from $20$ to $40 MHz$, corresponding to a range resolution of around $7$ to $3.5 meters$, which is not sufficient to distinguish between different parts within the body range (normally less than $2meters$). Moreover, independently considering positions may lead to the discontinuity in reconstructed movements. On the other hand, the $\mu$-Doppler spectrogram is directly related to the velocity profiles: the position of a Doppler bin refers to the direction of a target movement and how fast it moves; the strength of a Doppler bin is determined by radar cross section, distance between the target and receivers, the speed overlay and other complex factors. Therefore, learning velocities is more consistent with the characteristic of $\mu$-Doppler spectrograms. After obtaining velocities, we use a simple update equation to build the time dependency between poses, as the following: 
\begin{dmath}
\label{eq: update}
    p_t=p_{t-1}+v_{t}\delta t
\end{dmath}
where $p$ and $v$ are pose and velocity vectors, their subscripts represent frame indices and $\delta t$ is the time interval between two frames. 

\subsection{Neural Network}
\label{subsec: velocity NN}
\begin{figure*}
\centering
\includegraphics[width=\linewidth]{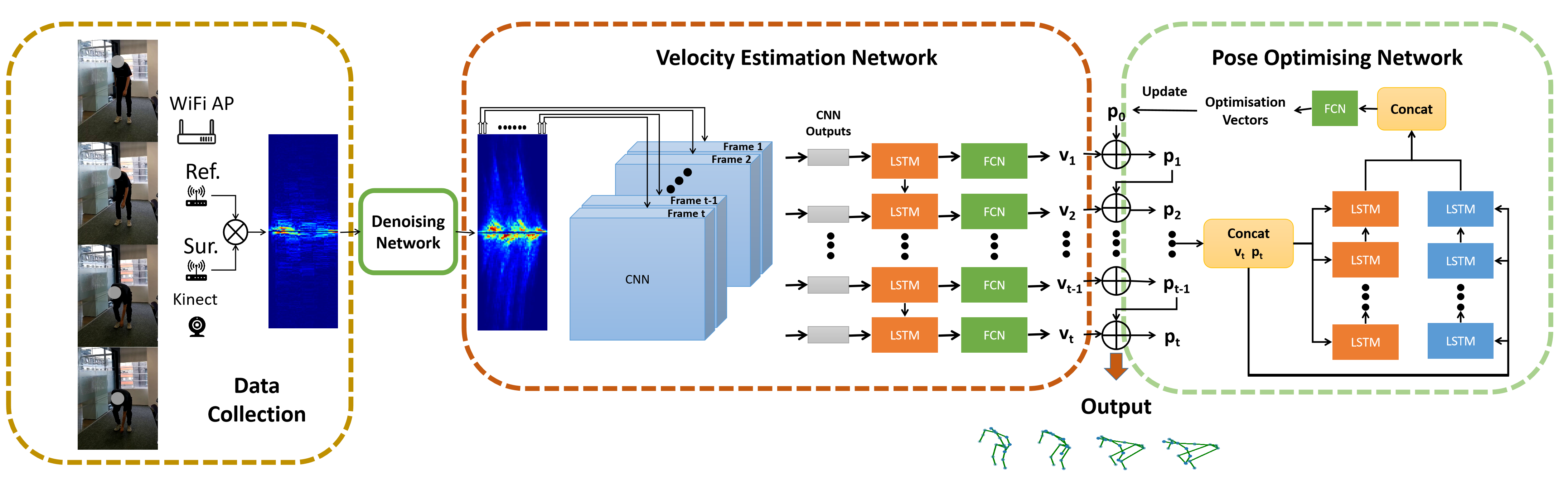}
\caption{The architecture details of MDPose}
\label{fig: MDPose}
\end{figure*}
To learn velocity sequence from a spectrogram, we develop a CNN-RNN architecture, as presented in the velocity estimation network in Fig. \ref{fig: MDPose}. We first use 1-dimensional convolutional layers over each frame to extract spatial features. These features are then concatenated along time axis and fed into a two-layer Long Short-Term Memory (LSTM) RNN to learn their time dependency. Finally, the outputs of each time steps are passed through fully-connected layers (FCs) to generate the velocity sequence. 

More specifically, we set the kernel size of convolutional layers to $5\times 5$ for the fine-grained feature extraction. Each convolutional layer is followed by a batch normalisation and rectified linear unit to accelerate training speed and add non-linearity to outputs of this layer. The final layer does not use any non-linear activation function and directly outputs estimated velocity values.

For training the network, we aim to minimize the difference between estimated velocities and the ground-truth velocities for each frame and key point. Therefore, the loss function can be defined as:
\begin{dmath}
    Loss=\frac{1}{T}\frac{1}{N}\sum_{t=1}^{T}\sum_{i=1}^{N}|v_{t}^{i'}-v_{t}^{i}|
\end{dmath}
where $T$ is the length of frames, $N$ is the number of key points, $v_{t}^{i'}$ and $v_{t}^{i}$ are estimated and real 3D velocity of $i^{th}$ key point at frame $t$.

\subsection{Training Data}
\label{subsec: Vest training data}
As a data-hungry algorithm, training a deep neural network often requires large volumes of data to retain strong generalisation performance. However, collecting experimental data for radar sensing systems is time-consuming and laborious. In Section \ref{sec: Noise}, we propose to use FMNet to remove environmental and multipath noise from measurement spectrograms. Furthermore, due to the denoised data is simulation-like, it is feasible to only use simulation spectrograms as the training data, and then apply the trained model to the denoised spectrograms. We have demonstrated in \cite{tang2021fmnet} significant classification improvement in practical applications by using this training strategy. Furthermore, we can easily produce a large amount of simulated data, so this method can greatly ease issues caused by the insufficient measured data. Meanwhile, to the best of the authors' knowledge, this is the first implementation of this approach.

\section{Pose Optimising Mechanism}
\label{sec: self-optimization}
Once we have extracted velocities from $\mu$-Doppler spectrograms, new questions arise: what is the initial pose $p_{0}$ for deriving the subsequent poses and how can we limit the accumulation of positional errors over time? For the first question, due to the restrictions of WiFi bandwidth, it cannot provide sufficient range resolution to determine positions of different body parts directly from the radar returns. However, a faulty $p_{0}$ may result in a sequence of illogical postural shifts. This can provide us clues about whether the current $p_{0}$ needs to be optimised. Furthermore, if we can make the correction periodically during a long-term motion reconstruction, we can effectively avoid the accumulation of positional errors. In this section, a bi-directional LSTM-based pose optimising mechanism will be introduced.

\subsection{Optimisation Vector}
\label{subsec: opt vec}
Although conventional supervised learning methods have ability to directly predict $p_{0}$ from $\mu$-Doppler or velocity profiles. it may lead to over-fitting and requires a significant amount of training data to maintain good generalisation performance. Due to the uncertainty of $p_{0}$, we can first initialise it based on some empirical guesses, such as a person standing before sitting, or a universal initialisation like T-pose, indicated as $p_{0}^{'}$. The challenge then becomes determining a vector $p_{diff}$ such that $P_{0}=P_{0}^{'}+P_{diff}$. But again, because of possible over-fitting issue, directly learning $p_{diff}$ is risky. Therefore, we decompose the problem and gradually approach the optimal $p_{0}$ by learning optimisation vectors ($\hat{OV}$) that is defined as:
\begin{dmath}
    \hat{OV_{i}}=\frac{\overrightarrow{p_{0}^{'}(i)p_{0}(i)}}{||\overrightarrow{p_{0}^{'}(i)p_{0}(i)}||}
\end{dmath}
where $||*||$ is the norm of a vector. This vector can let us optimise $p_{0}^{'}$ by moving it with a short distance in the appropriate direction, and enable us to approach the optimal position by repeating this process. This method has a high fault-tolerance and can effectively prevent over-fitting problems.
\subsection{Learn Optimisation Vector}
\label{subsec: training data}
Given $V=[v_1,v_2,...,v_t]$ and an initial pose, the pose sequence $P=[p_1,p_2,...,p_t]$ can be calculated with Equation \ref{eq: update}. As discussed at the beginning, this sequence might be illogical because of a erroneous $p_{0}$ but it could provide clues about how to find $\hat{OV}$. Therefore, we construct a optimising model that includes a bi-directional LSTM and FCs to estimate $\hat{OV}$ based on $V$ and $P$, as shown in the pose optimising network in Fig. \ref{fig: MDPose}.

For creating training dataset, we randomly select a state from the entire Mocap measurement as $p_{0}^{'}$ and generate a $P$ sequence. Then we define the input features of the network as the element-wise concatenation of $V$ and $P$, 
$\begin{bmatrix}
v_1 & ... & v_t\\
p_1 & ... & p_t
\end{bmatrix}$
, while the corresponding label is $\hat{OV}$. Furthermore, the objective function is defined as the following:
\begin{dmath}
    \min Loss=\frac{1}{N}\sum_{i=1}^{N}(1-cos(\frac{\hat{OV_{i}}^{'}\cdot \hat{OV_{i}}}{||\hat{OV_{i}}^{'}||||\hat{OV_{i}}||})) \\+\frac{1}{N}\sum_{i=1}^{N}(1-||\hat{OV_{i}}^{'}||)^2
\end{dmath}
where $[\cdot]$ refers to dot product, $\hat{OV_{i}}^{'}$ is the predicted vector for the $i^{th}$ joint. For the first term, we aims to minimize the angle between the $\hat{OV_{i}}^{'}$ and $\hat{OV_{i}}$, and the second term is used to limit the length of the vector. 

In practice, we can select numerous $p_{0}^{'}$ for the same velocity sequence to generate more training data, which can highly enhance the model's generalisation performance.
\subsection{Algorithm Overview}
\label{subsec: algorithm overview}
Based on this network, the pose optimising mechanism is proposed as shown in Algorithm \ref{alg: pose optim}. We also define a hyper-parameter, optimization rate (optr), to adjust the step of each optimisation, which is $0.01$ in our experiments. 
\begin{algorithm}
\caption{Pose optimising mechanism}\label{alg: pose optim}
\begin{algorithmic}
\Require the previous initial pose, $p_{0}^{'-}$; velocity sequence, $V$
\While{not converged}
\State $P \gets V$ and $p_{0}^{'-}$
\State Network input $\gets$ concatenate $P$ and $V$
\State $\hat{OV}^{'} \gets$ optimising network
\State Updated initial pose $p_{0}^{'+} \gets p_{0}^{'-}+\hat{OV}^{'} \times optr$
\EndWhile
\end{algorithmic}
\end{algorithm}

Apart from the initialization of the first frame, we can also apply this mechanism periodically during a long-term skeletal reconstruction. In this case, rather than beginning from a random pose, we will use the pose from the previous frame as a starting point and optimise it, which can significantly lessen the effect due to positional error accumulation.
\section{Data Collection and Experimental Details}
\label{sec: data collection}
\subsection{Dataset}
\label{subsec: dataset}
To demonstrate the performance of MDPose, we carried out various experiments to collect data from $7$ subjects and $3$ different environments. In the dataset, activities include walking towards the receiver (W+), walking away the receiver (W-), turn-around (TR), sit-down (SD), stand-up (SU), aggressive hitting (HT), passive covering (CV), pick-up (PU) and body rotation (BR). We also recorded some combined movements, such as W+$\rightarrow$TR$\rightarrow$SD and SU$\rightarrow$W-$\rightarrow$PU. Each is completed within $5$ to $10$ seconds and the length of the whole measurement is around $190$ minutes, resulting in over $2000$ individual activities (also called $2000$ samples). Furthermore, we deployed a Kinect sensor to simultaneously collect the ground-truth Mocap data for labelling the measurement data and generating simulation data. Therefore, every measured spectrograms has associated matched velocity profiles and simulated spectrograms. Some data examples are shown in Fig. \ref{fig: vels}.

We can see that the dataset contains various activities, such as movements with continuous changes (W+, W-), quick and strenuous movements (HT, CV), movements with the similar $\mu$-Doppler pattern (SD, HT), etc. This can provide comprehensive scenarios to validate the robustness of the framework. Furthermore, when we split the data into training and testing sets, apart from common splitting strategies i.e. splitting the whole dataset with $70\%$/$60\%$/$50\%$ split rates, we divided the dataset with the split rate of $23\%$. This training set only has samples from one person and one environment with the length of around $44$ minutes, while the testing set contains data from different subjects and different environments. This extreme situation is to test whether MDPose still performs well with different subjects and in new environments.

\subsection{SimHumalator}
To generate the simulation spectrograms based on Mocap data, we used SimHumalator, which is an open source simulation tool that can produce human $\mu$-Doppler spectrograms for the PWR sensing scenarios. It can be downloaded for free from \url{https://https://uwsl.co.uk/} and more details about SimHumalator are presented in \cite{vishwakarma2021simhumalator}. During data generation phase, SimHumalator uses the same WiFi standard and PWR parameter settings as the real experiments to guarantee that the same motion information is conveyed in the measured and simulated data. 

\subsection{PWR Experimental Setup}
\label{subsec: experimental setup}
\begin{figure}
\centering
\includegraphics[width=\linewidth]{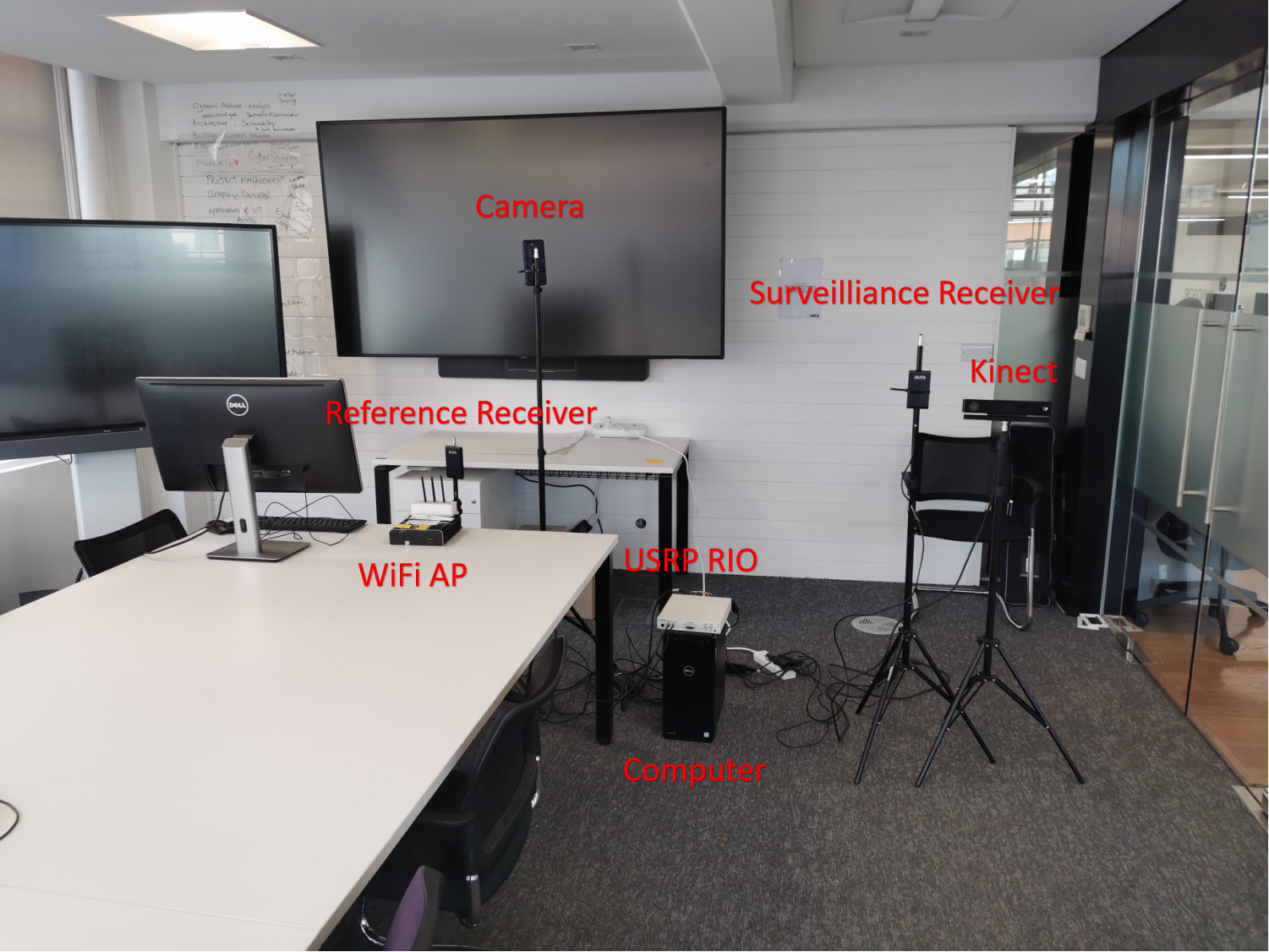}
\caption{The experimental setup of PWR system}
\label{fig: system setup}
\end{figure}
As shown in Fig. \ref{fig: system setup}, the PWR system uses one surveillance channel that is placed with the Kinect sensor together to simultaneously capture subjects' radar returns and velocity information, and uses one reference channel that is placed close to the WiFi AP to receive reference signals. We have implemented the PWR system in three different experimental environments: a broad living room, a narrow living room and a broad meeting room. In each case, the geometrical layout of the surveillance and reference channels are not identical but are relatively similar. This is to prevent having too much difference in the measurement data. On the other hand, a NI USRP RIO software-defined radio is connected to the two channels for the channel synchronization and real-time signal acquisition.

\subsection{Training Networks}
\begin{table*}[]
\resizebox{\textwidth}{!}{%
\begin{tabular}{|c|c|c|c|c|c|}
\hline
Network & Layer & Parameter Setting & BatchNorm & Activation & Training Setting \\ \hline
\multirow{7}{*}{Velocity Estimation} & Conv1d & filters 32, kernel size 5x5,  stride 2x2, padding 0 & Yes & ReLU & \multirow{7}{*}{\begin{tabular}[c]{@{}c@{}}optimizer: Adam\\ learning rate: 0,001\\ batch size: 64\end{tabular}} \\ \cline{2-5}
 & Conv1d & filters 64, kernel size 5x5,  stride 2x2, padding 0 & Yes & ReLU &  \\ \cline{2-5}
 & Conv1d & filters 64, kernel size 5x5,  stride 2x2, padding 0 & Yes & ReLU &  \\ \cline{2-5}
 & LSTM & hidden size 64, number of layers 2, bidirectional True & No & - &  \\ \cline{2-5}
 & Linear & input size 128, output size 128, bias True & Yes & ReLU &  \\ \cline{2-5}
 & Linear & input size 128, output size 51, bias True & Yes & ReLU &  \\ \cline{2-5}
 & Linear & input size 51, output size 51, bias True & No & - &  \\ \hline
\multirow{3}{*}{Optimisation Vector Estimation} & LSTM & hidden size 51, number of layers 2, bidirectional True & No & - & \multirow{3}{*}{\begin{tabular}[c]{@{}c@{}}optimizer: Adam\\ learning rate: 0,001\\ batch size: 128\end{tabular}} \\ \cline{2-5}
 & Linear & input size 102, output size 256, bias True & No & ReLU &  \\ \cline{2-5}
 & Linear & input size 256, output size 51, bias True & No & Tanh &  \\ \hline
\end{tabular}%
}
\caption{The architecture and training details of the two networks}
\label{tab:training}
\end{table*}
All neural networks are implemented with Pytorch\cite{paszke2017automatic} library on NVIDIA Quadro RTX 4000 graphics processing unit, and the architecture details and the training hyper-parameter settings of each networks have been listed in Table \ref{tab:training}.

\section{Experimental Results}
\label{sec: results}
In this section, a thorough evaluation of MDPose will be carried out both quantitatively and qualitatively. Comparisons with state-of-the-art approaches will also be made.
\subsection{Velocity Estimation}
\label{subsec: vest performance}
\begin{figure}
\centering
\includegraphics[width=\linewidth]{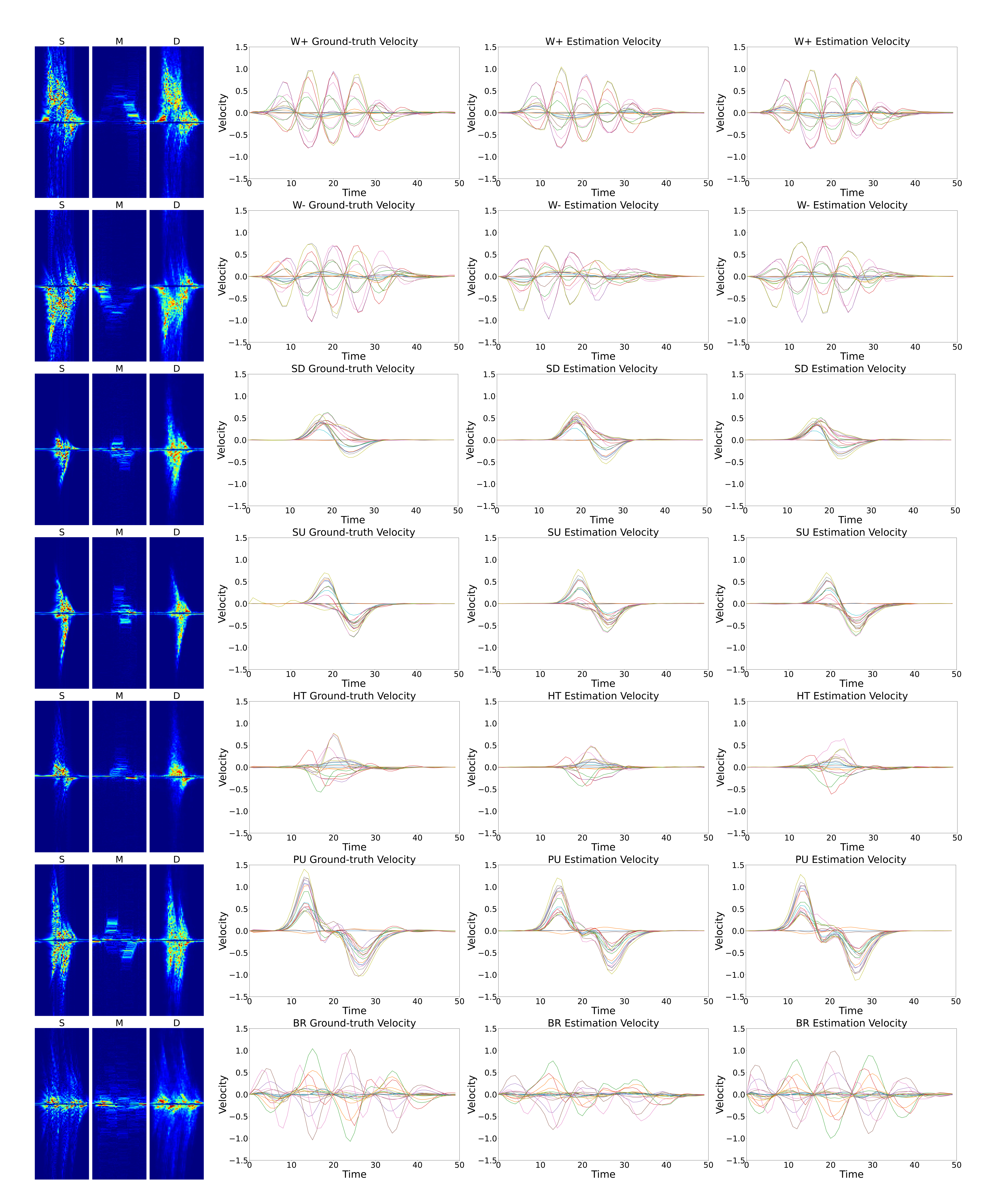}
\caption{The examples of velocity estimation results of different activities. From the left to right, they are simulation spectrogram, measurement spectrogram, denoised spectrogram, ground-truth velocity plot for $17$ key points, MDPose (M)-estimated velocity plot and MDPose (D)-estimated velocity plot.}
\label{fig: vels}
\end{figure}

\begin{table*}[]
\resizebox{\textwidth}{!}{%
\begin{tabular}{|c|ccccccccccccccccc|c|}
\hline
Joints & 1 & 2 & 3 & 4 & 5 & 6 & 7 & 8 & 9 & 10 & 11 & 12 & 13 & 14 & 15 & 16 & 17 & Overall \\ \hline
M & 0 & 1.6 & 6.0 & 8.5 & 1.9 & 5.6 & 8 & 2.8 & 4.5 & 7.7 & 14.5 & 15.8 & 7.6 & 14.8 & 18.5 & 5.7 & 7.6 & 7.7 \\ \hline
D & 0 & 1.5 & 6.1 & 8.5 & 1.5 & 6.0 & 8.4 & 2.6 & 4.9 & 7.2 & 10.5 & 11.7 & 7 & 12.4 & 15.8 & 5.1 & 6.9 & 6.8 \\ \hline
\end{tabular}%
}
\caption{Velocity estimation errors of each key points (unit: mm/frame)}
\label{tab:vel err}
\end{table*}
As introduced in Section \ref{sec: Noise}, spectrograms will be first denoised with FMNet, which aims to remove unwanted interference and reconstruct clear motion details. Furthermore, it enables training the network with the simulation data. To demonstrate the advantages of this strategy, we compared its velocity estimation results with those obtained from networks trained directly with measurement data. We denote simulated, measured and denoised spectrograms as $S$, $M$ and $D$, respectively. 

In Fig. \ref{fig: vels}, the first column presents the matched $S$, $M$ and $D$ of different activities. From $M$s, we can observe the overall direction of movement as well as the magnitude of velocities of different body parts. For example, in W+, the subject approaches the receiver from a distance. Therefore, the overall magnitude of the velocity increases first and then decreases. Meanwhile, when the person gets closer, the power of the received signal rises, making the Doppler pattern more visible. Such properties are important for applications like localisation and ranging, however, when retrieving human motion, the weak signal might be masked by noise, resulting in information loss. In contrast, the effects of signal attenuation and noise are limited in $S$. So, we can observe a distinct and complete Doppler pattern, which is beneficial for the human skeletal reconstruction task. And this is also why we made denoising and data enhancement as the first step of MDPose. From $D$s, we can observe that FMNet can effectively clean $M$s and generate data that is close to corresponding $S$s. Although some local 
regions remain blurred, the enhanced features can significantly improve the performance of velocity estimates, which can be demonstrated fom velocity plots.

The second to fourth columns in Fig. \ref{fig: vels} are plots of the ground-truth velocity, the estimated velocity based on $M$ and the estimated velocity based on $D$, respectively. Overall, MDPose can successfully extract velocity information from both $M$ and $D$, and most results have consistent patterns with the ground-truth. However, we can see that the $D$-based estimation has better performance than the $M$-based estimation, most notably in the plots for W- and BR. We can observe that in the weak signal sections, the corresponding $M$-based estimations have considerably lower velocity amplitudes than they should, which is because the motion features are masked by the background noise and the model fails to extract them. On the other hand, after being processed by FMNet, $D$-based estimations do not have the same issue and all of them are qualitatively close to the ground-truth plots. Furthermore, since the D-based results are obtained using a CNN-RNN model trained on $S$, the difference between the $D$- and $M$-based performances might be more evident in the case of a limited training set.

For the quantitative analysis, we present the velocity errors of each key point in Table \ref{tab:vel err}. We compared the mean absolute errors between $M$- and $D$-based results. It shows that the $D$-based results have less errors than the $M$-based results for most of key points, and even with some exceptions ($3$, $6$, $7$, $9$), their errors are still at a low level. For some points, the $D$-based results can achieve maximum $4.1mm/frame$ (i.e. around $41mm/s$) improvement, and this difference would significantly affect the long-term motion estimation. Finally, the overall error of $D$-based MDPose is $6.8mm/frame$ which is $0.9mm/frame$ better than $M$-based MDPose.

To sum up, we have demonstrated that MDPose can extract velocity properties from $\mu$-Doppler spectrograms. Meanwhile, due to clearer features, the $D$-based estimation normally outperforms the $M$-based estimation. However, $D$ cannot completely restore the details in $S$, and there are still some ambiguities. In this case, we may further improve the model's robustness by adding noise to $S$ during the network training. On the other hand, we believe other kinds of denoising algorithms can also improve the estimation performance and will be the subject of future research.
\subsection{Pose Estimation}
\label{subsec: poseest results}
\begin{figure}[t]
\centering
\includegraphics[width=\linewidth]{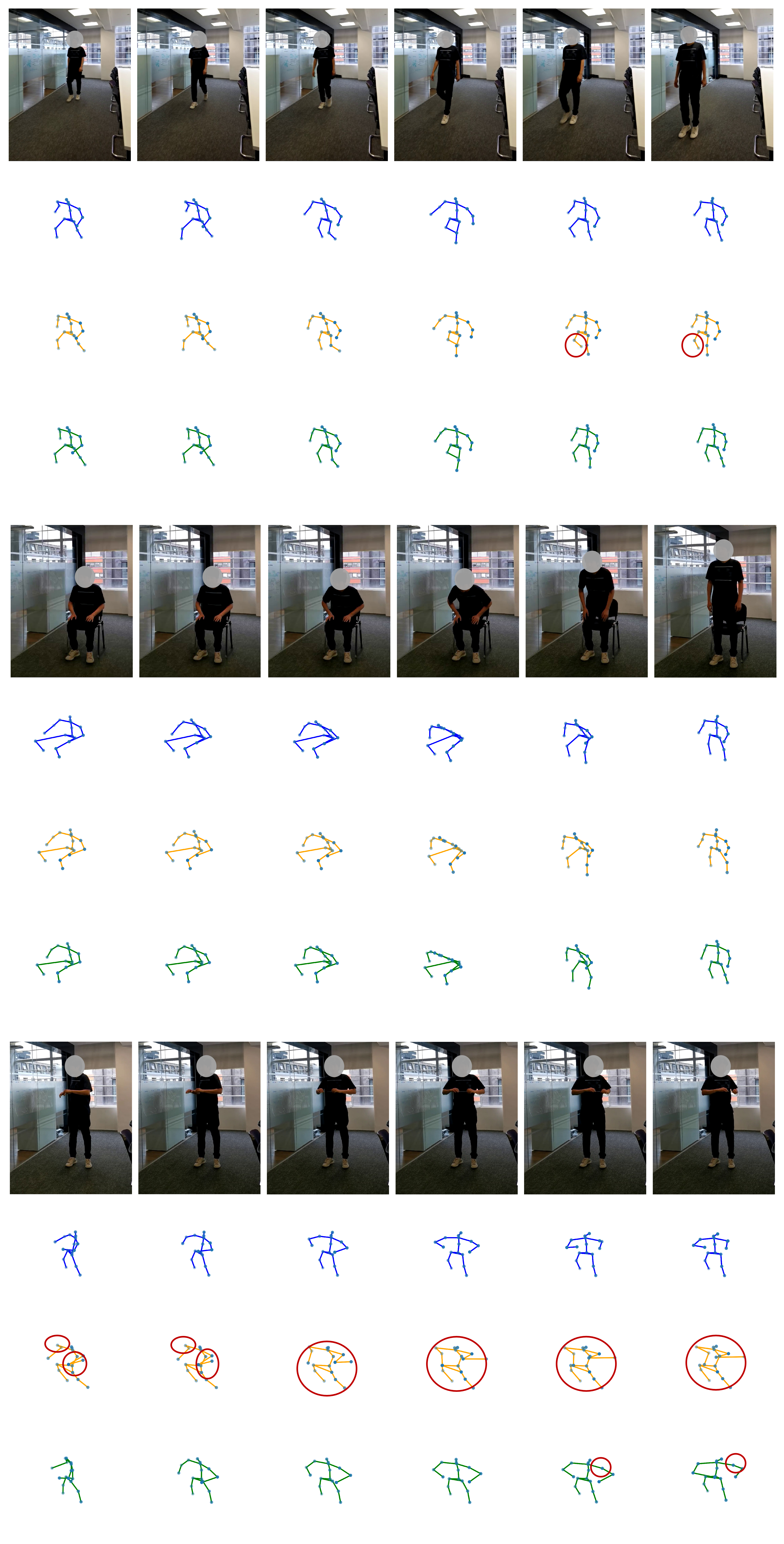}
\caption{The examples of pose estimation results of different activities: the first activity is W+, the second activity is SU and the third activity is BR; the blue skeleton is the ground-truth pose, the yellow one is $M$-based MDPose result and the green one is $D$-based MDPose result.}
\label{fig: pose}
\end{figure}

\begin{table*}[]
\centering
\resizebox{\linewidth}{!}{%
\begin{tabular}{|cc|cccccccccccclcccc|c|}
\hline
\multicolumn{2}{|c|}{Joints} & 1 & 2 & 3 & 4 & 5 & 6 & 7 & 8 & 9 & 10 & 11 & 12 & 13 & 14 & 15 & 16 & 17 & Overall \\ \hline
\multicolumn{1}{|c|}{\multirow{2}{*}{W+}} & M & 0 & 11 & 21 & 18 & 13 & 29 & 27 & 17 & 21 & 24 & 35 & 35 & 31 & 12 & 21 & 29 & 23 & 22 \\ \cline{2-2}
\multicolumn{1}{|c|}{} & D & 0 & 10 & 21 & 29 & 9 & 22 & 25 & 12 & 13 & 13 & 23 & 15 & 23 & 20 & 11 & 20 & 31 & 17 \\ \hline
\multicolumn{1}{|c|}{\multirow{2}{*}{W-}} & M & 0 & 14 & 24 & 31 & 11 & 25 & 36 & 13 & 14 & 30 & 29 & 30 & 32 & 46 & 46 & 24 & 28 & 23 \\ \cline{2-2}
\multicolumn{1}{|c|}{} & D & 0 & 12 & 16 & 31 & 6 & 23 & 36 & 13 & 13 & 14 & 24 & 21 & 28 & 26 & 42 & 20 & 25 & 21 \\ \hline
\multicolumn{1}{|c|}{\multirow{2}{*}{SD}} & M & 0 & 10 & 54 & 54 & 61 & 28 & 45 & 61 & 25 & 26 & 32 & 50 & 33 & 33 & 33 & 30 & 30 & 34 \\ \cline{2-2}
\multicolumn{1}{|c|}{} & D & 0 & 5 & 32 & 35 & 9 & 3 & 53 & 26 & 28 & 24 & 38 & 41 & 22 & 33 & 38 & 30 & 26 & 26 \\ \hline
\multicolumn{1}{|c|}{\multirow{2}{*}{SU}} & M & 0 & 13 & 24 & 28 & 8 & 25 & 24 & 23 & 25 & 25 & 27 & 27 & 27 & 21 & 20 & 28 & 25 & 22 \\ \cline{2-2}
\multicolumn{1}{|c|}{} & D & 0 & 4 & 28 & 24 & 17 & 27 & 20 & 16 & 18 & 13 & 22 & 25 & 19 & 17 & 16 & 22 & 16 & 18 \\ \hline
\multicolumn{1}{|c|}{\multirow{2}{*}{HT}} & M & 0 & 41 & 20 & 42 & 44 & 27 & 52 & 31 & 25 & 47 & 44 & 133 & 95 & 41 & 118 & 18 & 27 & 47 \\ \cline{2-2}
\multicolumn{1}{|c|}{} & D & 0 & 42 & 20 & 28 & 48 & 25 & 39 & 19 & 20 & 25 & 32 & 57 & 16 & 31 & 113 & 19 & 20 & 17 \\ \hline
\multicolumn{1}{|c|}{\multirow{2}{*}{CV}} & M & 0 & 8 & 70 & 120 & 13 & 57 & 91 & 20 & 37 & 37 & 53 & 158 & 281 & 146 & 237 & 39 & 65 & 84 \\ \cline{2-2}
\multicolumn{1}{|c|}{} & D & 0 & 20 & 74 & 117 & 12 & 46 & 83 & 24 & 35 & 43 & 52 & 71 & 60 & 46 & 82 & 37 & 59 & 51 \\ \hline
\multicolumn{1}{|c|}{\multirow{2}{*}{PU}} & M & 0 & 15 & 31 & 35 & 22 & 27 & 26 & 32 & 39 & 39 & 40 & 45 & 39 & 43 & 44 & 47 & 53 & 52 \\ \cline{2-2}
\multicolumn{1}{|c|}{} & D & 0 & 21 & 30 & 25 & 11 & 29 & 26 & 21 & 31 & 30 & 33 & 45 & 41 & 49 & 50 & 40 & 40 & 31 \\ \hline
\multicolumn{1}{|c|}{\multirow{2}{*}{BR}} & M & 0 & 63 & 54 & 36 & 71 & 29 & 52 & 30 & 26 & 42 & 92 & 147 & 112 & 153 & 205 & 37 & 57 & 71 \\ \cline{2-2}
\multicolumn{1}{|c|}{} & D & 0 & 21 & 16 & 17 & 15 & 25 & 19 & 11 & 22 & 25 & 80 & 85 & 42 & 92 & 100 & 29 & 36 & 37 \\ \hline
\multicolumn{1}{|c|}{\multirow{2}{*}{Overall}} & M & 0 & 22 & 37 & 46 & 30 & 31 & 44 & 28 & 27 & 34 & 44 & 78 & 81 & 62 & 91 & 32 & 39 & 44 \\ \cline{2-2}
\multicolumn{1}{|c|}{} & D & 0 & 17 & 29 & 38 & 16 & 25 & 41 & 18 & 23 & 23 & 38 & 45 & 31 & 39 & 57 & 27 & 32 & 29 \\ \hline
\end{tabular}%
}
\caption{Pose estimation error Comparison (unit: mm)}
\label{tab:pos err}
\end{table*}

\begin{table}[]
\centering
\begin{tabular}{|c|c|c|c|}
\hline
RFPose3D (CSI) & WiPose (CSI) & MDPose (M) & MDPose (D) \\ \hline
43.6 & 28.3 & 44.3 & 29.4 \\ \hline
\end{tabular}
\caption{Pose reconstruction error comparison with the state-of-the-art results}
\label{tab:comparison}
\end{table}
After obtaining the velocity sequence, we can use the pose optimising method to initialise the start pose and calculate the subsequent sequence of poses.. We presented some frames in Fig. \ref{fig: pose} for the qualitative analysis.

Video frames are illustrated on the first row which are followed by the matched ground-truth poses (blue), $M$-based poses (yellow) and $D$-based poses (green). Meanwhile, we have circled the areas where the error is greater than $100mm$. We observe that MDPose can accurately calculate the pose sequence from the velocity and initial pose estimations. For $D$-based results, most of them are consistent with the ground-truth poses, and matched with the real video recordings. Even if there are relatively large errors in some frames, they do not affect the overall performance and are still within acceptable limits. Moreover, comparing $M$- and $D$-based results, we can observe that the failure of the velocity estimation due to the low quality $\mu$-Doppler spectrogram significantly affects the pose estimation, especially in the third activity BR. This once again illustrates the importance of denoising and feature enhancement before feeding spectrograms into the network. Additionally, we can observe a phenomena: since the subsequent sequences are derived using the prior pose and velocity, large errors in the earlier pose are likely to be inherited by the future sequences, resulting in a continuous accumulation of errors, such as the last two frames of $M$-based results in W+, the last two frames of $D$-based results in BR and all frames of $M$-based results in BR. This, therefore, will have an impact on MDPose performance in terms of long-term skeletal motion estimation. In this case, we can address this issue with the optimising mechanism, which will be discussed in Section \ref{subsec: pose optimising}.

Apart from the qualitative results, we also presented the quantitative analysis in Table \ref{tab:pos err}. It thoroughly compares pose errors of $M$- and $D$-based MDPose results over different activities and different key points. Again, the $D$-based results are superior compared to the $M$-based results in most cases. Particularly for CV and BR, key points $13$ and $15$ of $M$-based results have more than $200mm$ errors, which have a significant impact on skeletal reconstruction. By contrast, in these cases, the $D$-based estimations maintain the low level of errors. Furthermore, by comparing overall errors of different activities and key points, all of the $D$-based results have less errors than the $M$-based results, with the mean absolute error of around $29mm$, while it is around $44mm$ for the $M$-based results. 

Furthermore, we also compared MDPose results with the state-of-the-art results, as shown in Table \ref{tab:comparison}. We can observe that MDPose ($D$) reduces the error by around $14.2mm$ when compared to RFPose3D (CSI), considerably improving the reconstruction accuracy. However, the MDPose (D) still performs slightly worse than the WiPose (CSI), with a difference of $1.1mm$. This is because the current single surveillance-channel setup of the MDPose does not compare favourably with the multi-receiver WiPose. With the addition of multiple channels in the future, we believe there is great potential to improve the performance of the MDPose. Additionally, although the performance of MDPose (M) is somewhat worse relative to the other results, it can still achieve comparable performance to RFPose3D (CSI), which could also indicate the effectiveness of the MDPose framework.

\subsection{Pose Optimising}
\label{subsec: pose optimising}
\begin{figure}
\centering
\includegraphics[width=\linewidth]{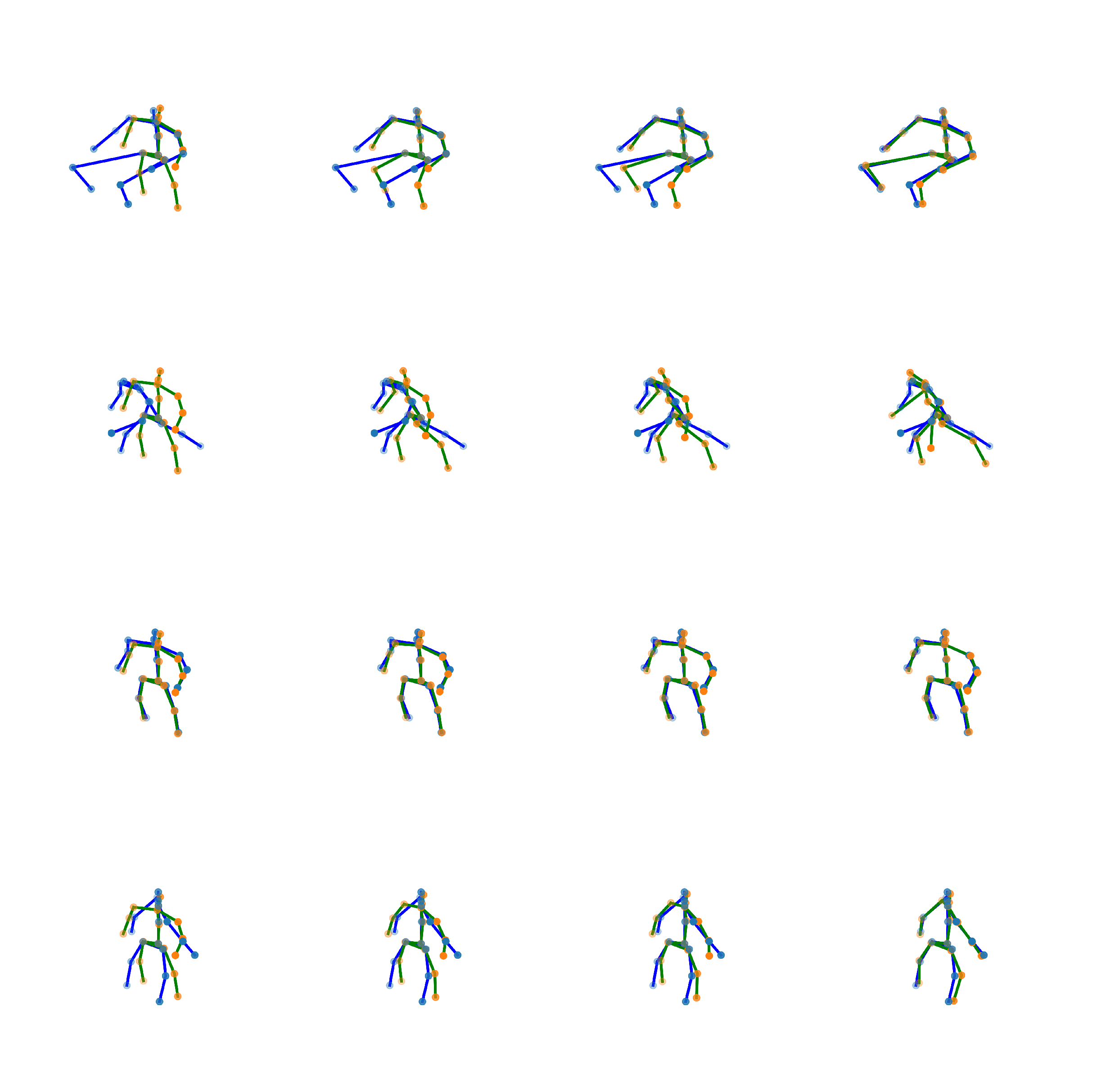}
\caption{The examples of pose optimising mechanism for initialising $p_0$}
\label{fig: pose_opt}
\end{figure}

\begin{figure}
\centering
\includegraphics[width=\linewidth]{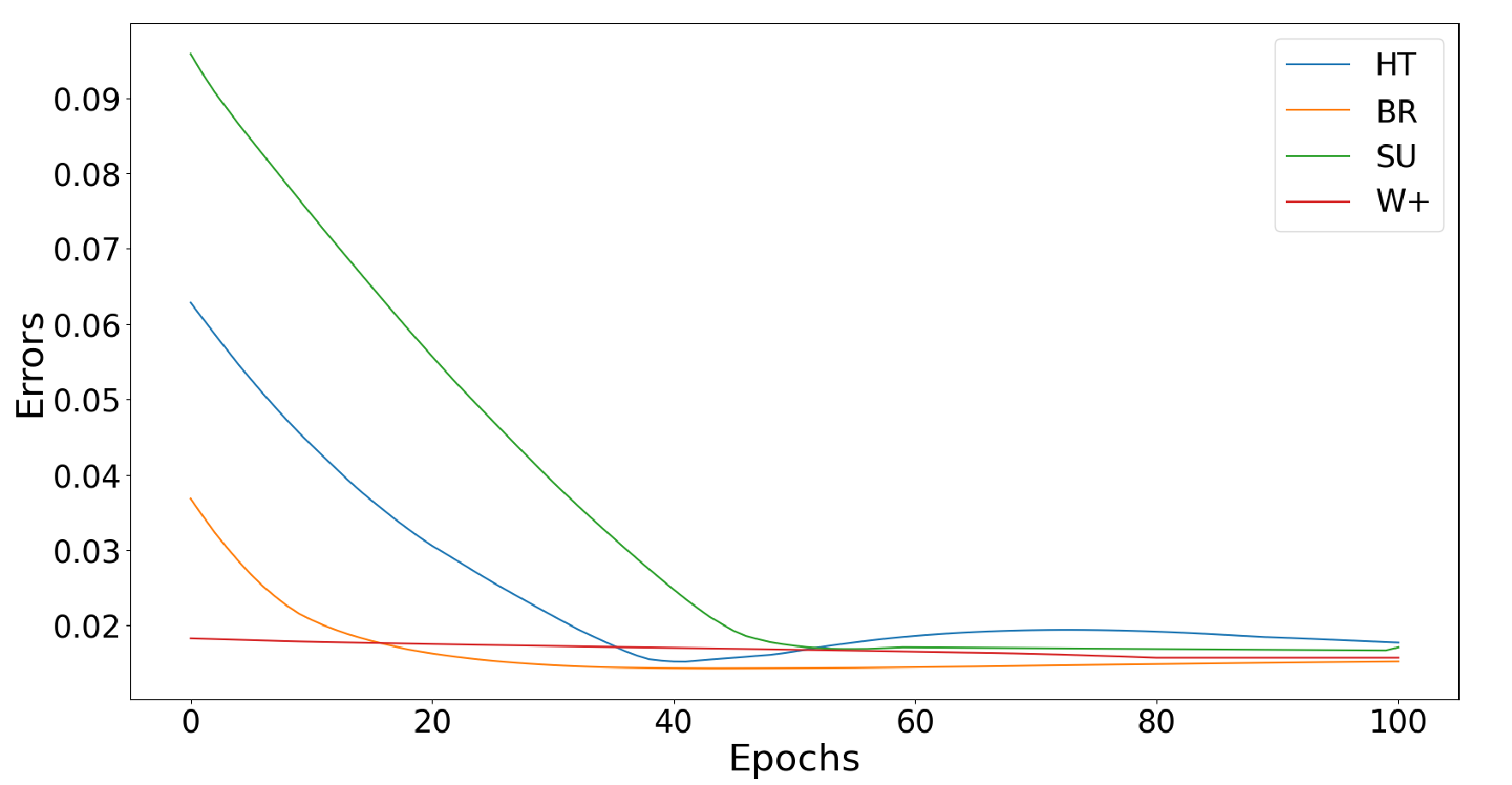}
\caption{The position error changes with the increase of optimisation epochs (y-axis unit: m)}
\label{fig: opt_err}
\end{figure}

\begin{figure}
\centering
\includegraphics[width=\linewidth]{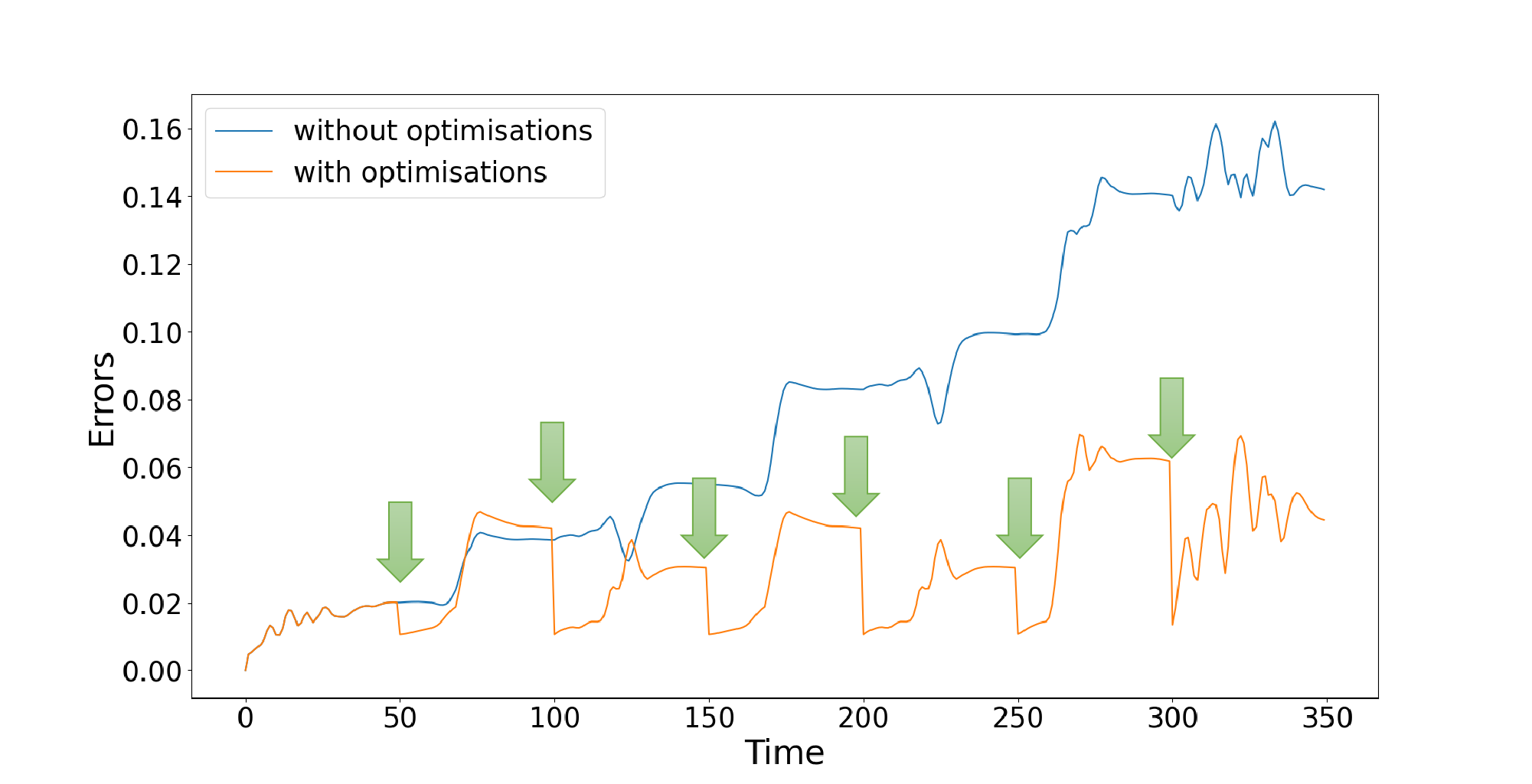}
\caption{The examples of pose optimising mechanism for optimising a long-term estimation: green arrows represent when we apply the optimising iteration}
\label{fig: opt_long}
\end{figure}
In our work, the pose optimising mechanism has two uses: initial pose estimation and error reduction in a long-term estimation.

For the initial pose estimation, we first randomly initialize a pose as the frame $0$, then the optimising mechanism can help it gradually approach the ideal starting pose, and finally it can be used to calculate pose sequence with the estimated velocities. We presented four examples in Fig. \ref{fig: pose_opt}. The blue skeleton with blue nodes is the ground-truth initial pose while the green skeleton with orange nodes is the estimated initial pose of each optimisation epoch. In practice, we used the same pose as the initialisation, as shown in the first column in Fig. \ref{fig: pose_opt}. We totally have $50$ optimisation epochs and the rest of columns present the results after processing every $10$ epochs. For the first row, the activity is SU, so the ideal initial pose is sitting-down. Therefore, when we initialise it with standing-up pose, the leg parts have large difference. However, as the optimisation proceeds, we can observe the legs progressively being adjusted to the proper positions. At the same time, the arms have also been slightly optimised to a better position. For the remaining points, they reach stability in the early epochs and are not affected by the other points. We observe similar characteristics in the rest of activities. Furthermore, for the third activity W+, the initialisation is quite close to the ground-truth pose. We can see that the optimisation algorithm still has a good handle in this case, and each point remains stable after a slight adjustment.

We quantified the mean absolute error between the estimated initial attitude and the ground truth initial attitude for the above activities, as shown in Fig. \ref{fig: opt_err}. From the plot, we can see that the errors start at relatively high values and gradually decrease with the optimisation process. Most of them can eventually be less than $20mm$.

On the other hand, the optimising mechansim can also be used in a long-term skeletal motion estimation to reduce the effect of the accumulative error. To test its performance, we collected a long $\mu$-Doppler spectrogram with around $350$ frames (i.e. 35 seconds). The sequence of activity is $SU\rightarrow W+\rightarrow PU\rightarrow BR\rightarrow W-\rightarrow SD\rightarrow SU$. From the blue line in Fig. \ref{fig: opt_long}, we can observe that without optimisation, previous errors can continually affect later estimations, resulting in the errors becoming increasingly larger. This phenomenon matches the discussion in Section \label{subsec: poseest results} and Fig. \ref{fig: pose}. By contrast, after applying optimising mechanism every $50$ frames as pointed by green arrows, the error can be periodically reduced so that the overall error fluctuates within an acceptable range.

According to above analysis, we have demonstrated that the proposed optimising mechanism is an effective method, which is not only an essential step in the skeletal reconstruction, but also plays an important role in further enhancing the quality of the estimations. Also by introducing the concept of $\hat{OV}$, the difficulty of training the model and the risk of overfitting are greatly reduced.
\subsection{Other Evaluation}
\label{subsec: other}
\begin{figure*}
\centering
\includegraphics[width=\linewidth]{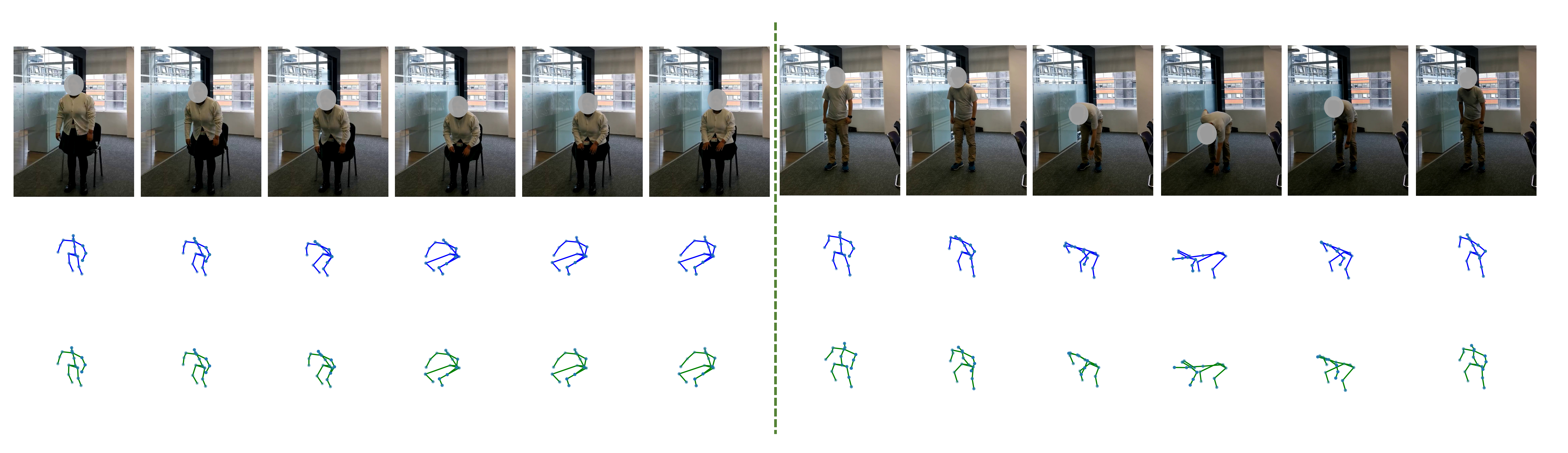}
\caption{The examples of pose estimation results of different subjects}
\label{fig: pose_otherppl}
\end{figure*}
\textbf{Generalisation Performance:} to validate the generalisation ability of MDPose under different indoor sensing scenarios, we have three experimental environments and $7$ subjects. We presented some examples of two different subjects in Fig. \ref{fig: pose_otherppl}. As we can observed, the MDPose can still successfully estimate poses of different subjects, and the results are consistent with the video frames and the ground-truth poses.

\begin{table}
\centering
\begin{tabular}{|c|c|c|c|c|}
\hline
Training Rate & 23\% & 50\% & 60\% & 70\% \\ \hline
RFPose3D (CSI) & - & 52.4 & 48.6 & 43.6 \\ \hline
WiPose (CSI) & - & 34.6 & 32.1 & 28.3 \\ \hline
MDPose (D) & 57.1 & 40.2 & 35.7 & 29.4 \\ \hline
\end{tabular}%
\caption{Pose reconstruction comparison with different training rates}
\label{tab:trainrate}
\end{table}
\textbf{Training Rate:} we also demonstrated how the error changes with the decrease of the training rate and compared MDPose ($D$) with the state-of-the-art results. We can see that the mean absolute error increases as the amount of training data decreases in all frameworks. Among them, MDPose outperforms RFPose3D (CSI) in all cases and is comparable to WiPose (CSI). In general, MDPose errors are maintained to a low level with a little impact on skeletal reconstruction task. Meanwhile, this result already can provide adequate and intuitive motion information to WiFi-based sensing scenarios. Furthermore, we also tested the robustness of MDPose by using only one subject's data for training and using other subjects' data for testing, and finally obtained the mean absolute error of $57.1mm$. This result is close to the error of RFPose3D (CSI) at a training rate of $50\%$, and is still an acceptable value for most scenarios, which once again validates the good generalisation performance and robustness of MDPose.

\begin{table}[t]
\centering
\begin{tabular}{|c|ccc|c|}
\hline
Frameworks & Denoising & CNN-RNN & Optimisation & Total \\ \hline
RFPose3D & - & - & - & 0.029 \\ \hline
WiPose & - & - & - & 0.016 \\ \hline
MDPose & 0.001 & 0.003 & 0.03 & 0.034 \\ \hline
\end{tabular}
\caption{Runtime analysis (unit: second)}
\label{tab:runtime}
\end{table}
\textbf{Runtime:} Table \ref{tab:runtime} presents the quantitative analysis of the efficiency of MDPose. We used NVIDIA Quadro RTX 4000 graphics processing unit and $10$ frames (i.e. around 1 second) $\mu$-Doppler spectrogram to test the time taken by the different components of MDPose. As we can see, the denoising and velocity estimation only used around $0.004$ seconds to obtain velocity properties. However, due to the lack of the initial pose, we have to iteratively use the optimising mechanism to estimate it, resulting in around $0.03$ seconds run-time.. Therefore, compared with other methods, MDPose may spend more time to estimate poses. However, this result can still demonstrate that MDPose is an efficient method which has ability to efficiently process long-term $\mu$-Doppler information.

\section{Conclusion}
\label{sec: conclusion}
In this paper, we presented a novel WiFi-based human skeletal motion reconstruction framework, MDPose, to effectively extract human motion from $\mu$-Doppler information. It has two main phases to achieve the task: the CNN-RNN-based velocity estimation and initial pose estimation (or pose optimising mechanism). Additionally, we highly recommend using an appropriate denoising algorithm to remove interference and enhance Doppler features. Herein, we developed FMNet to clean up noisy spectrograms, and we believe that other effective denoising methods can also improve the performance of MDPose. From experimental results, we have demonstrated that each component of MDPose works well: the denoising network can provide clean features for the later steps; the velocity estimation network can effectively extract velocity properties of up to $17$ body key points; the pose optimising mechanism not only helps to initialise the first pose, but also reduces the impact of accumulative errors on long-term estimations. Overall, MDPose has achieved the state-of-the-art performance with the estimation error of $29.4mm$. 

For future development of the research, there is still a lot of room for improvement with MDPose. Currently, we only use the single surveillance channel to collect Doppler information, which is not sensitive to velocity that varies along a path parallel to the receiver. Therefore, in our experiments, we only focused on processing activities perpendicular to the receiver, but this has limitations for some realistic scenarios. To address this issue, our future plan is to use a multi-receiver system to more comprehensively capture the motion properties. Furthermore, we will also explore the multi-people detection in our future development.

\section*{Acknowledgments}
This work is part of the OPERA project funded by the UK Engineering and Physical Sciences Research Council (EPSRC), Grant No: EP/R018677/1. 

\balance

\bibliographystyle{IEEEtran}

\bibliography{refs.bib}

\end{document}